\def\year{2021}\relax
\documentclass[letterpaper]{article} 
\usepackage{aaai21}  
\usepackage{times}  
\usepackage{helvet} 
\usepackage{courier}  
\usepackage[hyphens]{url}  
\usepackage{graphicx} 
\usepackage{natbib}  
\usepackage{caption} 
\usepackage{algorithm}
\usepackage{algorithmic} 
\usepackage{subfigure}
\usepackage{makecell}
\title{Incremental Knowledge-Based Question Answering}

\author {
    Yongqi Li,\textsuperscript{\rm 1}
    Wenjie Li, \textsuperscript{\rm 1}
    Liqiang Nie \textsuperscript{\rm 2} \\
}
\affiliations {
    \textsuperscript{\rm 1} The Hong Kong Polytechnic University \\
    \textsuperscript{\rm 2} Shandong University\\
    liyongqi0@gmail.com, wenjie.li@polyu.edu.hk, nieliqiang@gmail.com
}

\begin{document}

\maketitle

\begin{abstract}
In the past years, Knowledge-Based Question Answering (KBQA), which aims to answer natural language questions using facts in a knowledge base, has been well developed. Existing approaches often assume a static knowledge base. However, the knowledge is evolving over time in the real world. If we directly apply a fine-tuning strategy on an evolving knowledge base, it will suffer from a serious catastrophic forgetting problem. In this paper, we propose a new incremental KBQA learning framework that can progressively expand learning capacity as humans do. Specifically, it comprises a margin-distilled loss and a collaborative exemplar selection method, to overcome the catastrophic forgetting problem by taking advantage of knowledge distillation. We reorganize the SimpleQuestion dataset to evaluate the proposed incremental learning solution to KBQA. The comprehensive experiments demonstrate its effectiveness and efficiency when working with the evolving knowledge base.
\end{abstract}
\section{Introduction}
Knowledge bases, such as DBpedia\footnote{\url{http://dbpedia.org}.}, Freebase\footnote{\url{https://developers.google.com/freebase/}.}, and WikiData\footnote{\url{https://www.wikidata.org}.}, contain a large amount of facts about the real world. These facts are commonly represented as triples and each triple is in the form of $(s, r, o)$, where $s$, $r$ and $o$ represent the subject, the relation and the object, respectively.
The task of Knowledge-Based Question Answering (KBQA) aims to answer the questions presented in natural language using the relevant facts available in a knowledge base. 
 
\begin{figure*}
  \includegraphics[width=\linewidth]{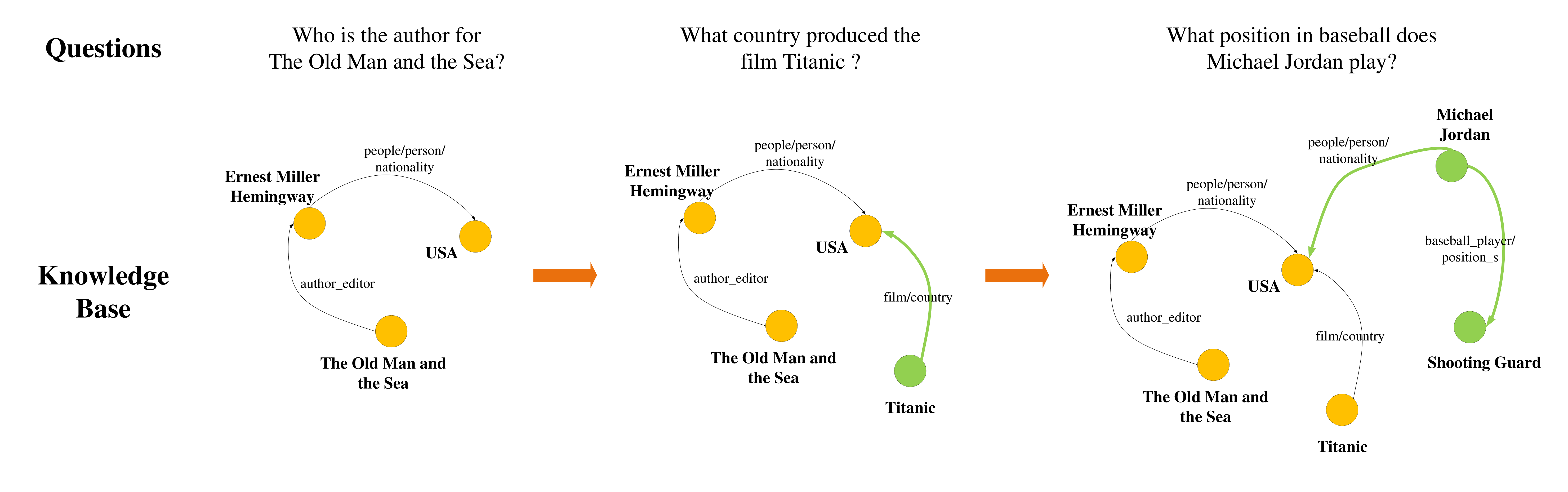}
  \caption{Illustration of the evolving knowledge base, where new relations and entities are supplemented constantly over time.}
  \label{example}
\end{figure*}
KBQA is an area well developed in the past years. The state-of-the-art research solution is to develop neural network-based models to learn the semantic similarity between a given question and the candidate facts in a knowledge base. A typical neural network-based KBQA model often involves two fundamental question analysis steps, i.e., entity linking and fact selection. First, the entity in the question is identified and linked it to its corresponding entities  in the knowledge base so that only a small subset of facts remains as candidates. Then, the answer to the question is extracted from the candidate fact that best matches the question. Essentially, the KBQA task is deemed as the {\it matching} problem with such models. 

The current KBQA models all assume the static knowledge base, i.e., once the knowledge is acquired, it no longer changes. However, in the real world, the knowledge base is evolving over time while the new entities and relations are constantly added into it\footnote{\label{ft:fool_note}For example, in the ICEWS05-15 dataset, which contains 461,329 events and their corresponding timestamps occurred during 2005-2015, there are in average about 3\% new types of relations and 14\% new entities every year.}. There have been some work exploring evolving KBs for link prediction and knowledge reasonin~\citep{garcia2018learning,trivedi2017know}. In question answering, when KB evolves, the questions related to the new knowledge should be readily answered, as illustrated in Figure~\ref{example}. However, it is difficult for a typical KBQA model to answer these questions because it has no ability to detect the relations that are not available in training ~\citep{wu2019learning}. To answer these questions, we can certainly re-train a new robust KBQA model over the entire data once the additional KB knowledge comes. Unfortunately, this is often impractical limited by the memory and computation resources, because it must rerun the model over the entire huge data even when the KB changes a little. Alternatively, it might be a better choice to fine-tune the parameters of the existing model with the new KB knowledge. However, when the model is fine-tuned using the new coming entities and relations alone, it will suffer from the serious problem of {\it catastrophic forgetting}~\citep{mccloskey1989catastrophic}. Since fine-turning focuses more on the new KB knowledge, when the model learns to apply the new knowledge, it will get to forget about what it has learned before and its ability to apply the old KB knowledge more or less diminishes. Therefore, how to develop a more intelligent QA model that can gradually expand its capacity and continually learn new knowledge while still preserving existing knowledge in maximum is a big challenge. To the best of my knowledge, this problem has not yet been explored in the KBQA area.

We cast the aforementioned problem caused by knowledge base evolvement into an incremental KBQA task. Incremental learning is critical to the problems where the data comes in a stream form~\citep{he2011incremental} and only a handful of them, called {\it exemplars} could be carried over to the subsequent time steps. Intuitively it is worth for us to draw lessons from it. Unfortunately, previous incremental learning approaches are mainly developed for classification~\citep{zhu2018merging} and it is infeasible to directly apply those successful incremental classification approaches to incremental KBQA. Recall that KBQA is essentially a matching problem. Building an incremental KBQA model is challenging because of the following reasons. First, compared with classification, KBQA is more complex. Given that questions, entities and relations all carry the semantic information, the entities/relations are often explicitly or implicitly inter-related in the organized knowledge base. Second, unlike classification which takes a single data sample as the input to decide the most probable class label, KBQA heavily relies on matching techniques and has to tackle multiple inputs including questions, entities and relations. Besides, incremental classification needs to grow the model parameters when new classes are added in, whereas KBQA does not. It is necessary to design a new incremental learning framework for KBQA. Third, since incremental KBQA is a new task, it is essential to design an appropriate dataset to evaluate the incremental learning abilities of models.  

In this paper, we present a compressive study on the incremental learning problem in KBQA. Our contributions are summarized as follows:

1. We innovatively define a new task, namely incremental KBQA and design an incremental KBQA dataset to evaluate the continuous learning ability of models.

2. We propose an incremental KBQA framework, takes the advantage of knowledge distillation to preserve utmost already-obtained machine knowledge when learning new knowledge for the matching problem.

3. We propose a novel exemplar selection technique, called collaborative exemplar selection, which considers the semantic relationships among the relations in the knowledge base when selecting representative samples from previous time steps to assist the training of the incremental model.

4. The experiments demonstrate that our proposed framework works well with the evolving knowledge base and alleviates the catastrophic forgetting problem effectively. And we will release the dataset and codes of this work.

 \begin{figure*}[t]
  \includegraphics[width=\linewidth]{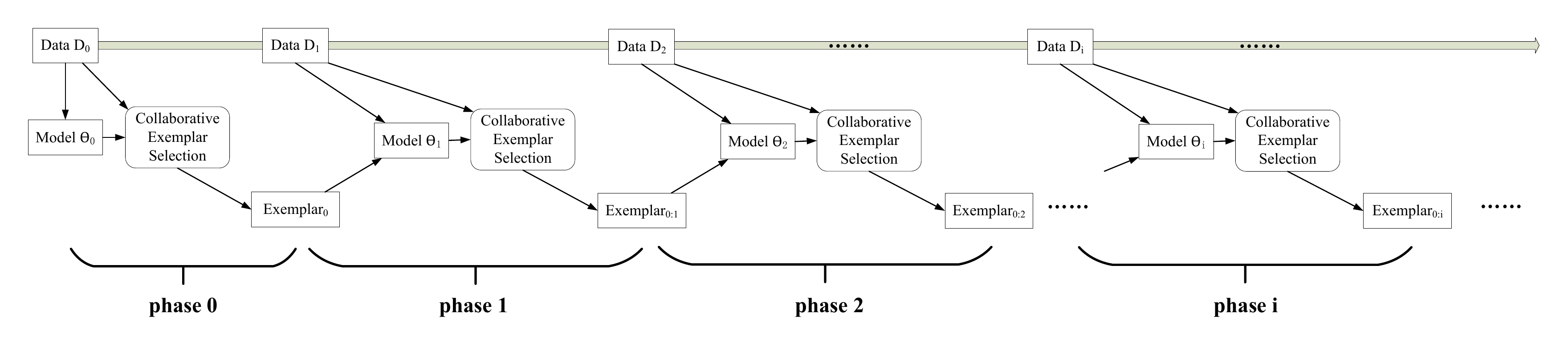}
  \caption{Illustration of our proposed incremental KBQA learning flow.}
  \label{framework}
\end{figure*}

\section{Related Work}
\subsection{KBQA}
The previous KBQA models can be roughly divided into two categories., i.e., semantic parsing based~\cite{yih2014semantic,reddy2014large,xu2018exploiting} and information retrieval based~\cite{bordes2015large,dong2015question,zhang2016question}.  To be more specific, the former models focus on mapping a question to its formal logical form to query on a KB. In contrast, the latter ones focus to measure the semantic similarity between the question text and the fact triples in a KB. Our work belongs to the latter line of study.

Recent years have witnessed the well development of information retrieval-based methods. For example, ~\citep{dong2015question} used multi-column convolutional neural networks (MCCNNs) to embed text without using any manual features and vocabularies. ~\citep{zhang2016question} introduced the attention mechanism to solve the matching problem at the character level. Recently, multi-hop question answering has attracted considerable attention~\cite{zhang2018variational,lin2018multi}, attaching importance to multi-relation reasoning and the structures of
knowledge bases. However, the previous models are all proposed to answer questions based on the static Knowledge base. In this work, we propose an incremental KBQA framework to learn from the evolving knowledge base. 
\subsection{Incremental Learning in Classification}
Incremental learning has a long history in machine learning~\cite{cauwenberghs2001incremental}.  Recent studies were generally conducted in the incremental-class setting, i.e., classes come in a sequence.  Grounded on the knowledge distillation technique~\cite{hinton2015distilling}, which was firstly applied in incremental learning by ~\citep{rebuffi2017icarl}, these studies tried to overcome the catastrophic forgetting problem via the distillation loss and exemplars. Several methods were proposed for exemplar selection, such as random~\citep{rebuffi2017icarl}, herding~\cite{castro2018end}, generation~\cite{shin2017continual} and end-to-end training~\cite{liu2020mnemonics}, etc. In our work, we consider inter-relationships among KB relations when selecting exemplars rather than only paying attention to the individual relations themselves. Moreover, different from the previous incremental learning approaches, which were specially proposed for classification, we design an incremental learning framework for matching-based KBQA.

Furthermore, there are some researches applying  incremental learning to NLP tasks. For example, ~\citet{shan2020learn} proposed a teacher-student model for text classification. \citet{wang2019incremental} studied incremental learning for task-oriented dialogue systems, where developers did not have to define user needs in advance. As for question answering, to the best of our knowledge, this is the first work to focus on the continual learning abilities of models in KBQA.

\section{Incremental KBQA}
\subsection{Problem Formulation}
We define the task of incremental KBQA as follows. Fact triples and the relevant questions are assumed to be incrementally available in sequence as the datasets $D_0, D_1, ...D_i, ...$, where $D_i=\{\mathcal{Q}_i,\mathcal{F}_i\}$, $\mathcal{Q}_i$ is the set of questions and $\mathcal{F}_i$ is the knowledge base represented by the fact triples $(s, r, o)$ at time $i$, where $s$, $r$ and $o$ represent the subject, the relation and the object respectively. The relations involved in the facts given in $D_i$ are completely new compared with the previously available ones. At time $i$, only $D_i$ and a handful of samples from previous time steps (called exemplars) can be used to train the KBQA model and the learned model is evaluated on all the test data from $D_1$ to $D_i$. The challenge for a successful incremental KBQA model is not only to learn the ability to effectively answer questions using the accumulated knowledge but also to learn from the dynamic streaming data efficiently.
\subsection{Overview}
The main workflow of the proposed incremental KBQA approach is illustrated in Figure~\ref{framework}. Assume there are $N$ phases, including one initial phase and $N-1$ incremental phases. The initial phase actually trains a KBQA model in a conventional way that feeds the positive and negative pairs of questions and fact triples in data $D_0$ to the model to learn the model parameter $\Theta_{0}$ (Section 3.3). In each subsequent incremental phase $i$, inspired by the work of incremental learning for classification~\cite{rebuffi2017icarl}, we feed not only the question and fact triple pairs in $D_i$ but also a number of exemplars, i.e., $Exemplar_{0:i-1}$, selected from previous phases  by applying the proposed collaborative exemplar selection strategy (Section 3.4). The incremental model in phase $i$ is then trained on data $D_i$ and $Exemplar_{0:i-1}$ collectively using a novel margin-distilled loss function (Section 3.5). 
\subsection{Incremental KBQA Learning}
Given data $D_i=\{\mathcal{Q}_i,\mathcal{F}_i\}$ and $Exemplar_{0:i-1}$, the model $\Theta_{i}$ is expected to learn answering questions with the new knowledge and at the meanwhile avoiding to lose the ability to handle the questions relevant to the old knowledge. In all incremental phases, entity linking is implemented to generate positive and negative pairs for training, and a deep neural network model is followed to encode the text inputs into embedding vectors. The gradient approach is used to update the weights of the network with the margin-distilled loss function.

\textbf{Entity Linking.} Similar to the previous work~\citep{dai2016cfo,wang2018apva}, we train a BiGRU-CRF model to detect the entity mentions in questions. Each question is then converted into a (mention, pattern) tuple, where the pattern is obtained by simply replacing the mention in the question with $<e>$. For each question, candidate fact triples are selected as follows. (1) We firstly apply the entity linking as previous methods\cite{he2016character,hao2018pattern}. In this work, we employ the Active Entity Linker, to obtain the top-20 entities for each question. Active Entity Linker is proposed in~\cite{yin2016simple} and the top-20 entities are publicly released \footnote{\url{https://github.com/Gorov/SimpleQuestions-EntityLinking}.}. It has been shown to be able to achieve better coverage of ground truth. The triples whose subjects are within the top-20 entities are selected as candidate fact triples. (2) For each of the top-20 entities, we randomly select some other relations~\cite{hsiao2017integrating} to generate new triples as candidate fact triples. And (3) some triples randomly selected from the knowledge bases are also included in the set of candidate fact triples. We build the candidate fact triples set from various aspects to be consistent with previous KBQA approaches for effective training and fair comparison. What is different from the previous approaches is that we enlarge the number of candidate fact triples in order to more reliably evaluate the model's ability to find the correct KB triple from a larger collection of candidates. 

Now, for each question, which has been converted into (mention, pattern), we combine it with its truly linked fact (subject, relation) as the positive pair, and the question (mention, pattern) with the candidate fact triples excluding the true one as negative pairs.  Besides, we denote the positive pair and negative pairs in $D_i$ as $pair_D^+$ and $pair_D^-$, and those pairs in $Exemplar_{0:i-1}$ as $pair_E^+$ and $pair_E^-$, respectively.

\textbf{Text Encoding.} We feed the texts of mention, pattern, subject, and relation into a deep neural network model to obtain their corresponding embedding vectors. This model plays the same role as the embedding layer in most conventional KBQA models. We choose the model presented in~\cite{yin2016simple}. Specifically, the mentions and subjects are encoded by a char-based CNN network. Considering they are short, the generated representations are more robust even in the presence of typos, spaces and other character violations via character-level rather than commonly-used word-level encoding. As for patterns and relations, a word-level CNN with attentive maxpooling is applied. These obtained embeddings are used to train the model using the proposed margin-distilled loss, which will be introduced later.
\subsection{Collaborative Exemplar Selection}
In the $i$-th phase, if the model can be retrained using all the relations and their corresponding questions accumulated up to this phase, i.e., {$D_0$, $D_1$, ... $D_i$}, the model $\Theta_i$ can learn to answer both the new and old types of questions well. However, when the amount of the data increases, the retraining will consume more and more computing resources. Due to memory limitation, retraining with the entire available data is often impossible. In practice, only a handful of samples are selected as $Exemplar_i$ from each phase $i$ to retrain the the model $\Theta_j$ $(j>i)$. Typically, the number of exemplars is set to be much smaller than that of the original data. As a result, how to select the most representative and effective exemplars always plays a very important role in incremental learning. 

The previous incremental learning approaches focus on the classification task generally select the samples that are near to the average vector (called herding) for per class. Considering the particularity of KBQA where the relations have their own text inputs, we propose a novel collaborative exemplar selection strategy based on the semantic information carried in texts. Different from the previous approaches that only consider the interior of one class, our approach considers semantic relationships among relations to collaboratively select the “best” learning samples from a global view. Inspired by the importance of support vectors in Support Vector Machine (SVM)~\cite{suykens1999least}, we assume that the samples near the boundary between relations contain more significant information for the model to detect the relation. For a relation, we consider the other relations semantically similar to it. As such, we are able to select the most effective samples for all the related relations collaboratively rather than only replying on the individual relations alone.

The exemplar selection procedure is specified in Algorithm~\ref{alg:CES}. For each relation $r1$ in $D_i$, we firstly feed its text to the encoder to obtain its embedding vector $V_{r1}$. Similarly, we obtain the embedding vector for each relation $r_2$ in $D_i$ and $Exemplar_{0:i-1}$. Then, we calculate the cosine similarities between the relation $r1$ and $r2$. We select the most $m$ similar relations.
Next, for each relation $r3$ from the $m$ most similar relations, we select $n$ questions corresponding to the relation $r1$ as exemplars. Specifically, we input all the patterns corresponding to the relation $r1$ to the model and obtain embedding vectors. We then calculate the cosine similarities between relation $r3$ and all the patterns. Furthermore, we select the $n$ most similar ones because we think these samples allow the model to distinguish similar relations. At the end $n$ samples for $m$ relations are selected.
\begin{algorithm}[tb] 
\caption{Collaborative Exemplar Selection.} 
\label{alg:CES} 

\begin{algorithmic}[1]
\REQUIRE ~~\\ 
  The input data in phase $i$, $D_i$;\\
  The model parameters, $\Theta_i$;\\
  The previous exemplars, $Exemplar_{0:i-1}$\\
  The number of considered relations, $m$;\\
  The number of samples selected for each considered relation, $n$;
\ENSURE ~~\\
The selected exemplars from $D_i$, $Exemplar_i$;
\STATE$Exemplar_i$=[]\\
\STATE for relation $r_1$ in $D_i$:\\
\STATE~~~~$V_{r1}$=Model($\Theta_i$, $r_1$)\\
\STATE~~~~s1=[]\\

\STATE~~~~for relation $r_2$ in $D_i$ $\cup$ $Exemplar_{0:i-1}$:\\
\STATE~~~~~~~~$V_{r2}$=Model($\Theta_i$, $r_2$)\\
\STATE~~~~~~~~s1.append(cos($V_{r1}$,$V_{r2}$))\\
\STATE~~~~s1=argmax(s1)[:m]\\
\STATE~~~~for relation $r_3$ in s1:\\
\STATE~~~~~~~~s2=[]\\
\STATE~~~~~~~~for pattern that corresponding to relation $r_1$:\\
\STATE~~~~~~~~~~~~$V_{pattern}$=Model($\Theta_i$, pattern)\\
\STATE~~~~~~~~~~~~s2.append(cos($V_{r3}$,$V_{pattern}$))\\
\STATE~~~~~~~~s2=argmax(s2)[:n]\\
\STATE~~~~~~~~$Exemplar_i$.add(s2)\\
\STATE RETURN $Exemplar_i$; 
\end{algorithmic}
\end{algorithm}

\subsection{Margin-Distilled Loss} 
Knowledge distillation is a technique initially proposed to transfer information between different neural networks or network structures. In this work, we apply it to a single model to distill knowledge between different time steps. It is expected that the knowledge in the model is retained as much as possible over time to overcome the {\it catastrophic forgetting} problem as mentioned before. 

We design a margin-distilled loss, a joint loss function combing a margin loss and a mean squared error (MSE) loss, especially for the KBQA matching problem. We feed the samples in the new data to the margin loss, which is same as those used in most previous KBQA approaches learning to rank the candidate fact triples. The margin loss term guarantees the model to learn the new KB knowledge. Meanwhile, we feed the samples in the exemplars to the MSE loss. This term is a distillation term that guides the model to retain the old KB knowledge across time steps.
In short, the margin loss is applied to the samples from $D_i$ while the MSE loss is used on the samples from $Exemplar_{0:i-1}$.

Specifically, for each pair, let’s denote the embedding vectors of the mention, the pattern, the subject and the relation obtained via text encoding as $V_{mention}$, $V_{pattern}$, $V_{subject}$, and $V_{relation}$, respectively. The matching similarity score is calculated as follows,
\begin{equation} \label{eqn1}
s_D^+ =cos(V_{mention},V_{subject})+cos(V_{pattern},V_{relation}),
\end{equation}
where $cos()$ is a cosine similarity function and the mention, the pattern, the subject and the relation are in $pair_D^+$. The similarity scores for $pair_D^-$, $pair_E^+$, and $pair_E^-$ computed by Eq.($\ref{eqn1}$) are denoted as $s_D^-$, $s_E^+$, and $s_E^-$, respectively.
Formally, the margin-distilled loss function is defined as follows: 
\begin{equation} \label{eqn2}
L=L_M + L_S,
\end{equation}
where $L_M$ is the margin loss and $L_S$ is the MSE loss. 

The margin loss $L_M$ is computed by
\begin{equation} \label{eqn3}
L_M=max(0, \lambda+s_D^- - s_D^+),
\end{equation}
where $\lambda$ is a hyper parameter, $s_D^+$ and $s_D^-$ are the similarity scores of the positive pair and negative pair in $D_i$, respectively. Based on the margin loss, the model learns to increase similarity scores of positive pairs and decrease similarity scores of negative pairs.
The MSE loss $L_S$ is formulated by,
\begin{equation} \label{eqn4}
L_s = (s_E^+ - l_E^+)^2 + (s_E^- - l_E^-)^2,
\end{equation}
where $s_E^+$ and $s_E^-$ are the similarity scores of the positive pair and negative pair in $Exemplar_{0:i-1}$, respectively. $l_E^+$ and $l_E^-$ are the labels calculated when samples are selected as exemplars in the preceding phases. For example, assume some samples have been selected as exemplars and the model $\Theta_j$ has been trained in phase $j$ ($j<i$). Then, for an exemplar, we obtain the positive and negative pairs via the entity linking step as mentioned before. Next, we load $\Theta_j$ and feed the positive and negative pairs into the model to obtain the embedding vectors. The two similarity scores of positive and negative pairs, i.e., $l_E^+$ and $l_E^-$ calculated by Eq.($\ref{eqn1}$) are deemed as target labels in the following phases.  We can see that the MSE loss allows the model to keep the previous scores from the old samples as much as possible. As a result, the model can still keep a fair performance on the old samples when it learns the new knowledge. In general, this is how we alleviate the catastrophic forgetting problem.

\section{Experiments and Evaluations}
\subsection{Dataet}
As claimed before, there is no appropriate QA dataset that contains the evolving knowledge and QA pairs for this new task. Following the previous work~\citep{rebuffi2017icarl,liu2020mnemonics} in incremental learning,
we re-organize the SimpleQuestion (SQ) Dataset~\cite{bordes2015large} to evaluate and compare the performance of KBQA models on the incremental setting. The SQ dataset consists of 108,442 single-relation questions and their corresponding fact triples $(s, r, o)$. 

We build the Incremental SimpleQuestion (Incremental-SQ) dataset from the SQ dataset as follows. (1) We firstly randomly shuffle and split the relation types into 5 sets $\mathcal{R}_i$, where $i=0$ to $4$. (2) Then we create 5 fact sets $\mathcal{F}_i=\{(s,r,o)\}$, where $r \in  \sum_{j=1}^{i} \mathcal{R}_j$. (3) Afterwards, we construct 5 question sets $\mathcal{Q}_i$, where these questions in $\mathcal{Q}_i$ match with the relations in $\mathcal{R}_i$. (4) Finally, we obtain 5 sub-datasets $D_i=(\mathcal{Q}_i, \mathcal{F}_i)$, where $i=0$ to $4$. It is guaranteed that for every question in $\mathcal{Q}_i$, its corresponding fact triple is in $\mathcal{F}_i$, and $\mathcal{F}_{i-1}$ is the subset of $\mathcal{F}_{i}$, which reflects the evolving knowledge base where the new relations types and entities are added successively. 
The statistics of the datasets is summarized in Table \ref{dataset statistics} . The reported experimental results in this paper are based on these datasets. Specifically, we use $80\%$ of questions in each $\mathcal{Q}_i$ to form the training set and the rest $10\%$ as the validation and test sets, respectively. It is worth mentioning that the number of the relations relevant to the questions in $\mathcal{Q}_i$ is different from that in $\mathcal{F}_{i}$, since not all relations in a knowledge base have the corresponding questions. 
\begin{table}[tbp]
\centering
\caption{Statistics of the dataset.}\label{dataset statistics}
 \scalebox{0.85}{
\begin{tabular}{c|ccccc}
\hline
Dataset& $D_0$ & $D_1$ & $D_2$ & $D_3$ & $D_4$\\ \hline
\makecell{\# q in train}& 18,001& 22,055&10,768&11,672&15,961\\
\makecell{\# q in validation}& 2,250&2,757&1,346&1,459&1,995\\
\makecell{\# q in test}& 2,251&2,757& 1,346&1,460&1,996\\
\makecell{\# relations in  $\mathcal{F}_{i}$}& 1696& 3,392&5,088&6,784&8,480\\
\# relations of q& 330& 305 &339 &329 &336\\
\hline
\end{tabular}
}
\end{table}

\begin{table*}[tbp]
\centering
\caption{Performance  comparison  between  our  proposed model and the previous KBQA methods. We calculate $Accuracy_i$ in each phase $i$ and $Average_a$ for all phases. }\label{Overall table}
 \scalebox{0.9}{
\begin{tabular}{|c|c|c|c|c|c|c|}
\hline
Approach& $Accuracy_0$ & $Accuracy_1$ & $Accuracy_2$ & $Accuracy_3$  & $Accuracy_4$ &$Average_a$ \\
\hline
Yin et al.& 0.8956 &0.8225 &0.7711 &0.7144 &0.6635 &0.7734               \\ \hline
Golub et al.& 0.8573 &0.71 &0.5211 &0.4792 &0.4559 &0.6047                    \\\hline
Hao et al.& 0.8764&0.8303 &0.7982 &0.7755 &0.7743 &0.8109                    \\\hline
Ours& 0.8956 &0.8648 &0.8396 &0.8317 &0.8312 &0.8525                  \\\hline
Ours w/o distillation loss& 0.8956 &0.8097&0.7950 &0.7871 &0.7639 & 0.8103                        \\\hline
Ours with random exemplars& 0.8956 &0.8389&0.8107 &0.8094 &0.8077 & 0.8324                      \\\hline
Ours with herding exemplars& 0.8956 &0.8453&0.8198 &0.8179 &0.8153 & 0.8388                     \\\hline
Upper bound& 0.8956 &0.8670&0.8478 &0.8494 &0.8486 & 0.8617                     \\\hline
\end{tabular}
}
\end{table*}

\subsection{Experimental Setups}
\textbf{Implementation Details.} In our approach, the dimensions of character embedding and word embedding are both set to 128. The CNN is equipped with 128*2 filters in two distinct sizes [2, 3]. The margin $\lambda$ in Eq.~\ref{eqn3} is 0.5. The number of relations $m$ used for selecting exemplars for each relation is 5 and $n$ is 2. We optimize the parameters using Adam with the initial learning rate 0.001, and the batch size is 256.\footnote{See Appendix A. for the details of the corresponding validation performance for each reported test result, the number of hyperparameter search trials and the expected validation performance.}.

\textbf{Evaluation Metric.} The evaluation metric is the accuracy. For each phase $i$, we calculate the accuracy $Accuracy_i$ as 
the number of the correctly answered questions divided by the total number of the questions, where the questions are those in $Test_0$ to $Test_i$, i.e., the test sets in $D_0$ to $D_i$. If a single number is preferable, we report the average of these $Accuracy_i$ as $Accuracy_a$  to reveal an average performance on all $N$ phases as follows:
\begin{equation} \label{eqn5}
Accuracy_a = \frac{\sum_{i=0}^{N-1} Accuracy_i }{N}.
\end{equation}

\begin{figure*}[t]
  \centering
  \subfigure[]{
  \includegraphics[width=0.32\linewidth]{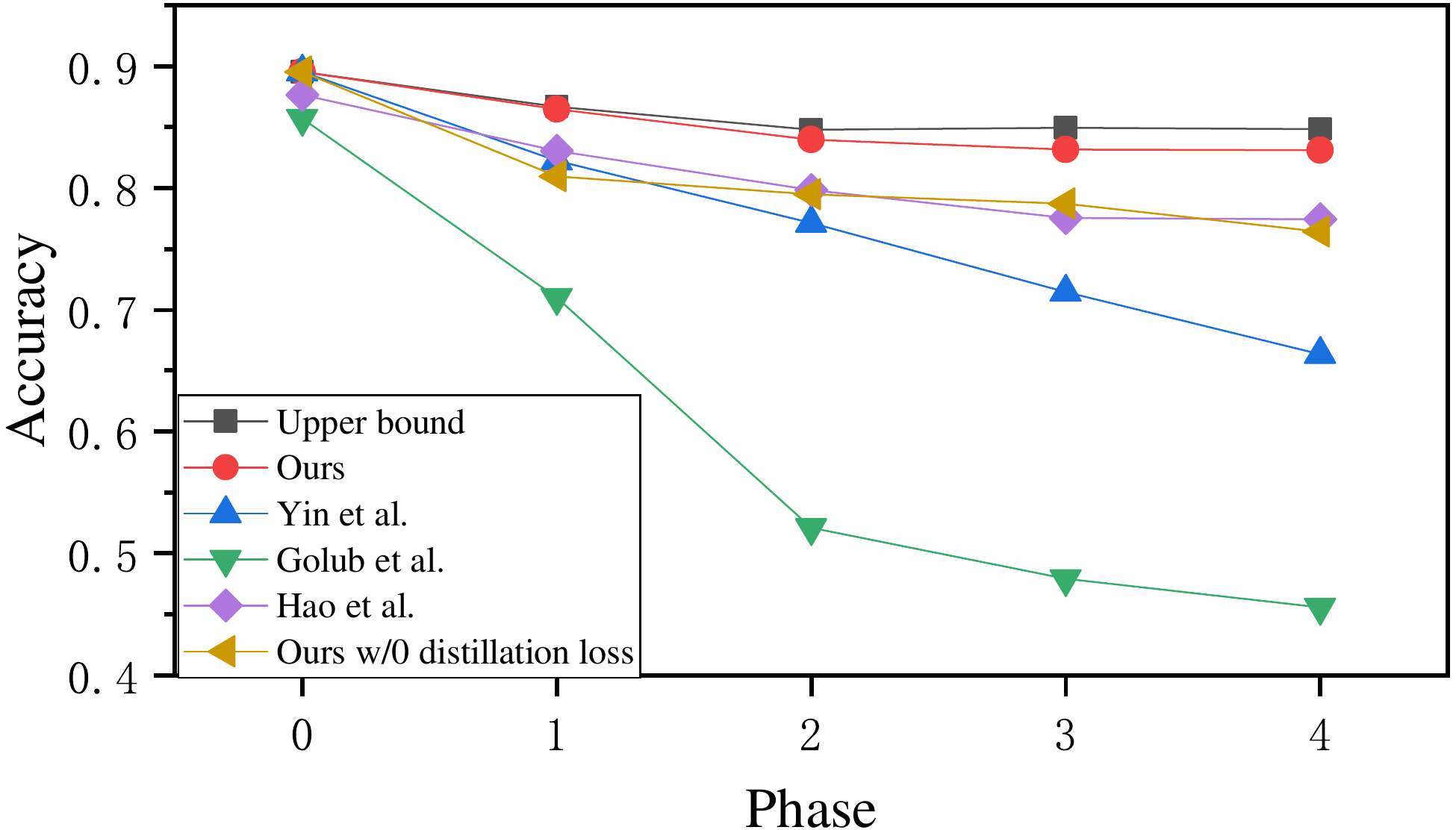}
   \label{fig5:subfig1}
   }
   \subfigure[]{
  \includegraphics[width=0.32\linewidth]{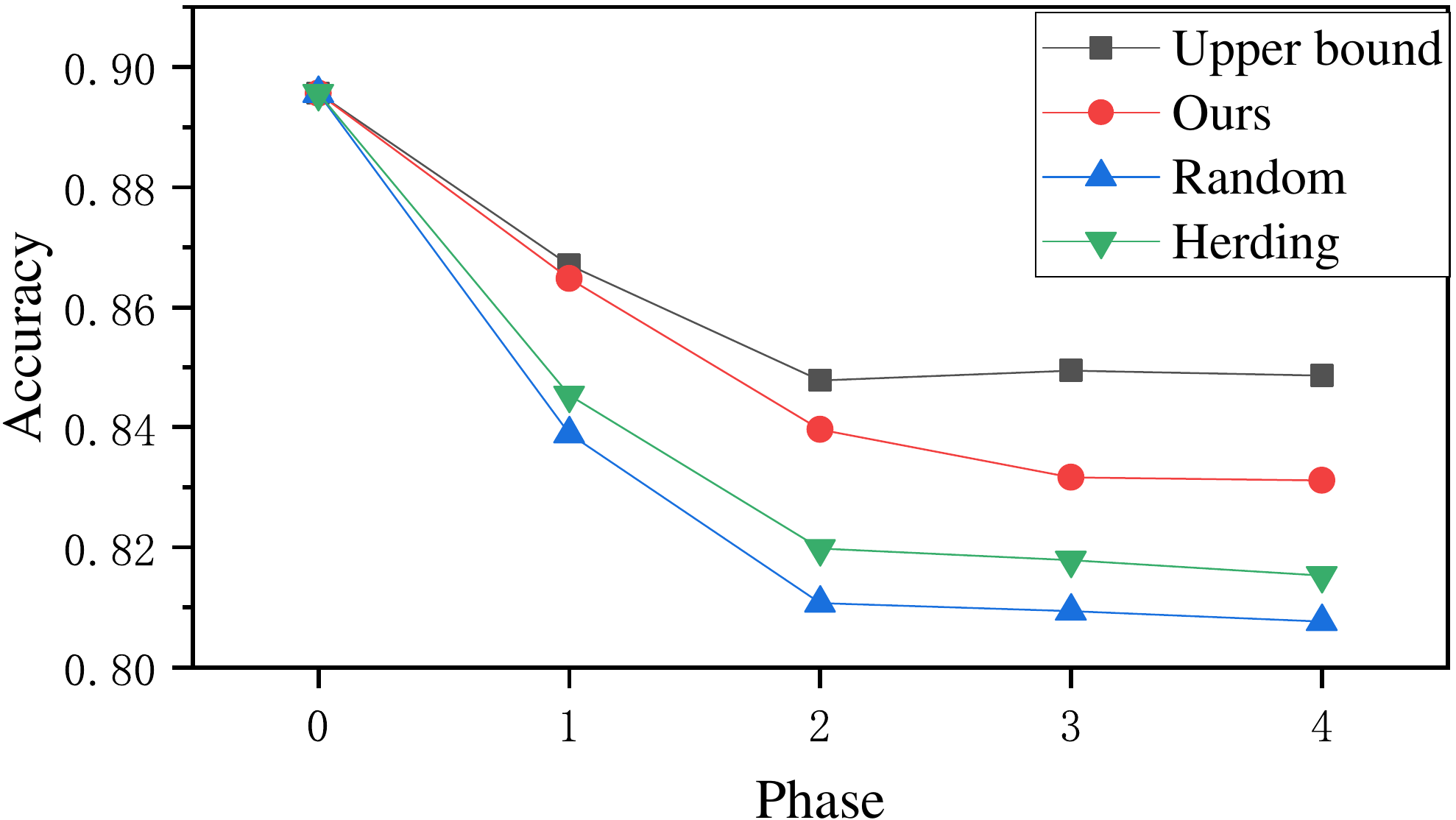}
   \label{fig5:subfig2}
   }
  \subfigure[]{
  \includegraphics[width=0.32\linewidth]{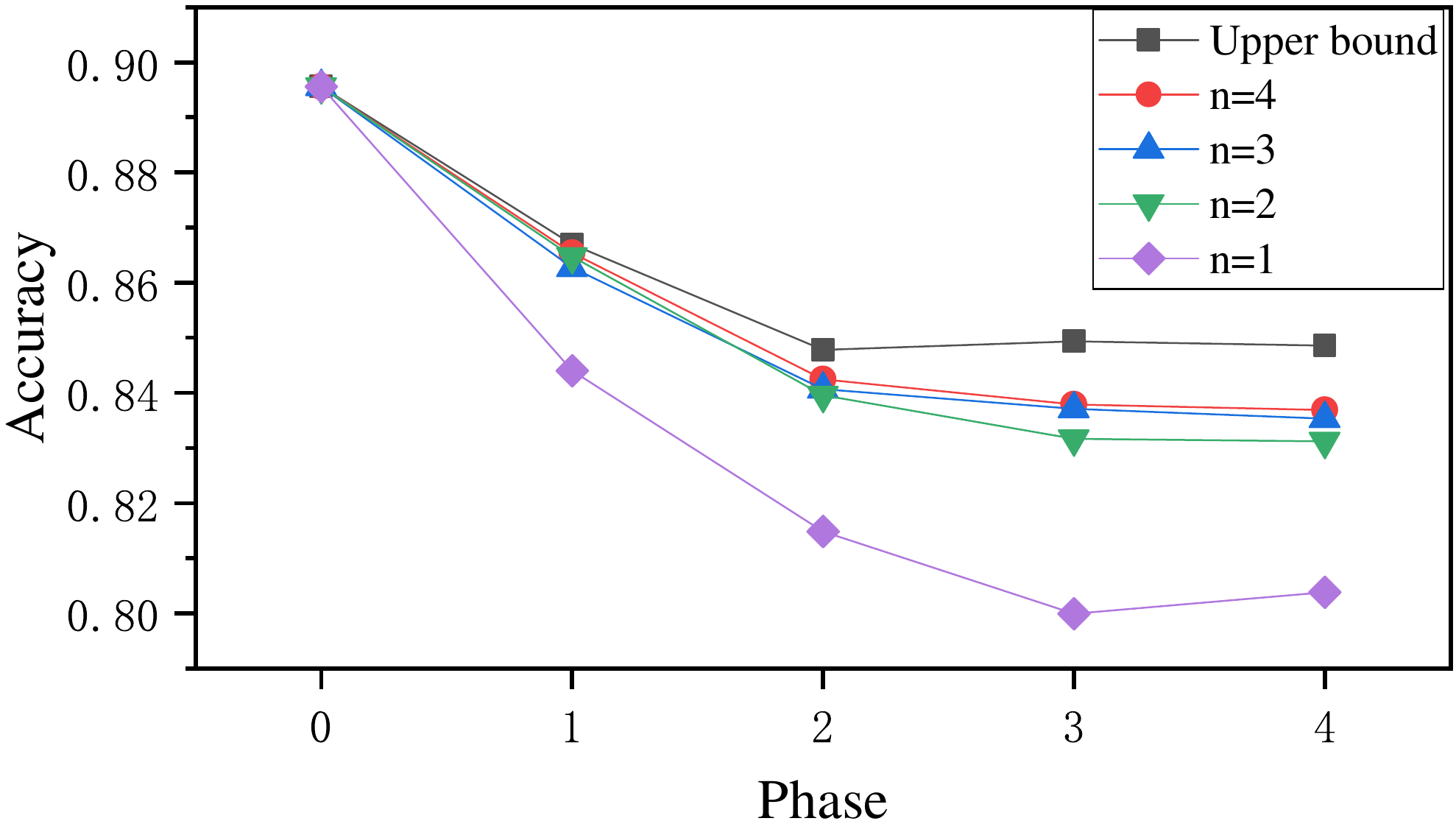}
   \label{fig5:subfig3}
   }
   \subfigure[]{
  \includegraphics[width=0.32\linewidth]{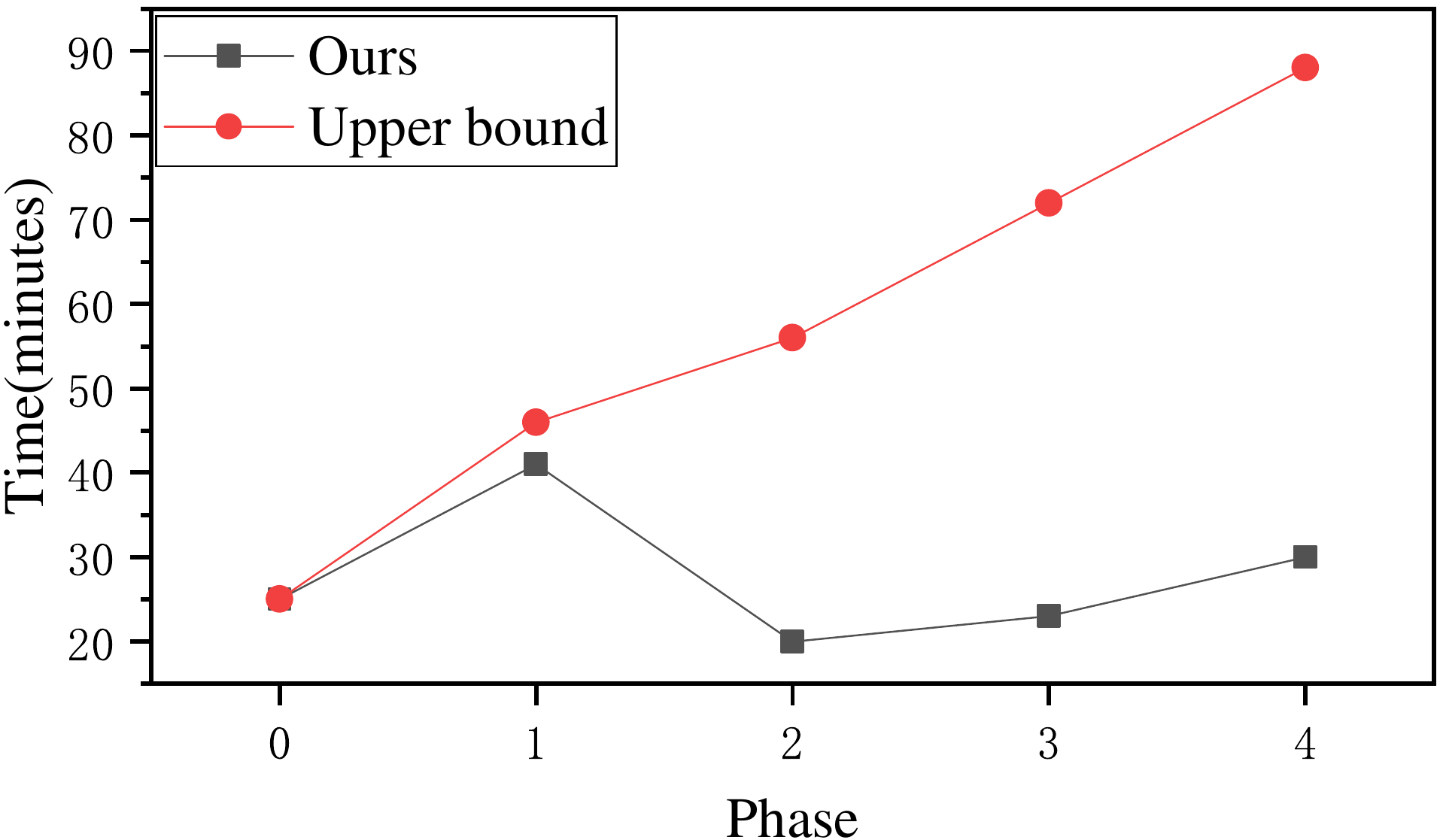}
   \label{fig5:subfig4}
   }
    \subfigure[]{
  \includegraphics[width=0.32\linewidth]{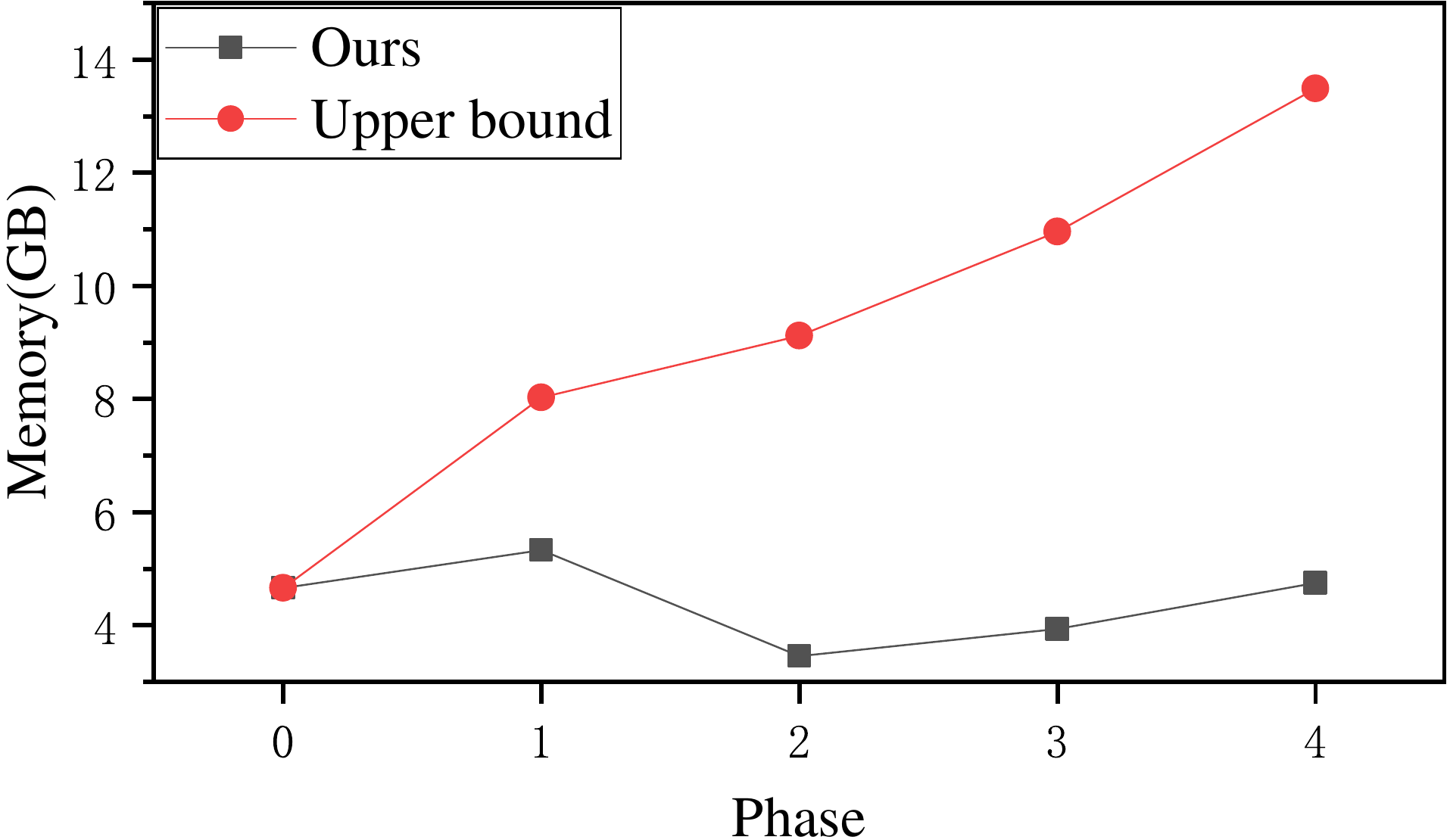}
   \label{fig5:subfig5}
   }

    \caption{ (a) Performance comparison between our proposed model and the previous KBQA methods. (b) Performance comparison between our proposed method and other exemplar selection methods; (c) Parameter analysis of the number of exemplars; and (d)-(e) the time and memory consuming between our method and the upper bound, respectively. }\label{fig5}
\end{figure*}

\subsection{Overall Performance}
To demonstrate the effectiveness of our proposed framework, we compare it with the following methods. ~\citet{yin2016simple} utilized an attentive CNN network to encode questions and fact triples at both word and character levels. ~\citet{he2016character} leveraged a character-level attention-based encoder-decoder LSTM model. ~\citet{hao2018pattern} added the pattern-revising procedure after entity linking to mitigate the error propagation problem. The performance of the above approaches is state-of-the-art on the SimpleQuestion dataset. As claimed by~\citet{mohammed2018strong}, more sophisticated models are unnecessary.
Since these models are developed for conventional KBQA, we fine-tune them with the new KB knowledge at each time step to suit for the incremental KBQA setup. Specifically, for phase $i$, we load the model parameters in the phase $i-1$ and continue training on the current $D_i$. We also report the upper bound performance, which retrains a new robust KBQA model over the entire data in each phase. The results of all models are summarized in Table~\ref{Overall table} and the curves of the $Accuracy_i$ for inspection are plotted in Figure~\ref{fig5:subfig1}, from which we obtain several observations as follows. 

(1) As the time step increases, the performances of all models decrease and are certainly below the upper bound. This clearly demonstrates the existence of the catastrophic forgetting problem mentioned before, i.e., the problem that the model forgets the old knowledge when it learns the new knowledge. The performance of the upper bound also declines in the first two incremental phases. That is because the results of entity linking in these two phases are worse.
(2) In Table~\ref{Overall table}, it is obvious that our model obtains the best average accuracy $Accuracy_a$ (excluding the upper bound), which verifies that the proposed framework is able to effectively alleviate the catastrophic forgetting problem and retains the evolving knowledge better than other models. In the initial phase, our model obtains the same accuracy as ~\citet{yin2016simple}, because we apply the same encoding method as it.  It is noticed that the model of ~\citet{he2016character} decreases the most, as this model completely replies on the character information. The size of the character vocabulary is quite small and thus the embeddings are unstable when training the model. 
(3) Compared with other methods, the performance of our method is very close to the upper bound. As claimed before, retraining is expensive because it needs to rerun the model over the huge data. We compare our method with the upper bound in terms of the time and memory consumption. The results are shown in Figure~\ref{fig5:subfig4} and Figure~\ref{fig5:subfig5}, respectively. We can see that as phases increases, the time and memory consumed by the upper bound increase linearly, which is because that the data is accumulated and expands linearly. As for our method, the time and memory are consistent with the number of data in each phase. There is an abrupt increase in phase $1$, because the data in phase $1$ is comparatively big as shown in Table \ref{dataset statistics}.  

\subsection{Component-wise Evaluation}
Next, we remove the distillation loss, i.e., $L_s$ in Eq.~\ref{eqn2}, to further investigate the effectiveness of our proposed margin-distilled loss. Specifically, we feed exemplars to the text encoder and then calculate the cosine similarities. Rather than putting the similarity scores into the distilled loss term as before, we put them into the margin loss $L_m$. This result is included in Figure~\ref{fig5:subfig2} and Table~\ref{Overall table} being ``Ours w/o distillation loss". By doing such a comparison, we gain the following insights. 

(1) Without the distillation loss, our model still performs better than Yin et al. although both use the same text encoder. It demonstrates that the exemplars themselves are useful to retain the old knowledge. It is expected that if the number of exemplars gets larger, the performance will be much better.
(2) we can see that "Ours w/o distillation loss" is worse than “Ours”, meaning that the distillation loss takes full advantages of exemplars to retain the old knowledge. Our framework works well because we rely on not only a handful of previous samples but also the margin-distilled loss, which actually plays an important role.

\subsection{Verification of Collaborative Exemplar Selection}
We further conduct the experiments to compare our proposed collaborative exemplar selection methods with some other exemplar selection alternatives. The simplest choice is to randomly select samples for per relation. The herding-based exemplar selection applied in \citet{rebuffi2017icarl} and \citet{castro2018end} selects the samples near the average embedding vector of a relation. We replace the exemplar selection method in our model with the above two, and the results are illustrated in Figure~\ref{fig5:subfig3}. It is found that herding-based selection surpasses random selection. This may be due to the fact that the samples around the centers are more representative. It can be also seen that our collaborative exemplar selection method exceeds the other two methods. Our method selects the samples that near the boundary among relations. It thus provides more effective information for the model to distinguish similar relations. To verify the robustness of our propose method, we test the statistical significance of the $Average_a$ difference between Ours and Ours with herding exemplars (the best among the compared methods). The p-value is smaller than 0.01.

To understand the influence of the number of exemplars, we conduct experiments by changing the number of samples $n$ selected for each considered relation in Algorithm~\ref{alg:CES}. The experimental results are illustrated in Figure~\ref{fig5:subfig3}, where we can see that the performance improves and approaches to the upper bound when $n$ increases.

\subsection{Catastrophic Forgetting Handling}
As analyzed before, one of the challenges in the incremental KBQA task is the catastrophic forgetting problem, i.e., the model may forget the old knowledge when learning the new KB knowledge. To reveal this problem, in each phase i, we report the accuracy on each available test set, i.e., $Test_0$ to $Test_i$, respectively. So, we can see the variation tendency on each test set as time steps increases.

We compare the method~\cite{yin2016simple} with ours for fair comparison, because we both use the same text encoder. Table~\ref{ Catastrophic Forgetting1} summarizes the detailed results of Yin et al. on each available test set in every phase.  It can be seen that the performance on each test set declines as time step increases. For example, the accuracy on $Test_0$ declines over phase 0, 1, 2 3, 4. Initially, the accuracy is 0.8956 and in the last phase it already decreases to 0.5852. This is because the knowledge learned in the first phase is being forgetting as time step increases. It again verifies the existence of the catastrophic forgetting problem. However, compared with Yin et al., our method performs much better, see Table~\ref{ Catastrophic Forgetting2}. For example, the accuracy on $Test_0$ declines over phase 0, 1, 2 3, 4. Initially, the accuracy is 0.8956 and it only drops about 0.02 at the end. Although the catastrophic forgetting problem still somewhat exists, our model can alleviate it well. 
\begin{table}[tbp]
\centering
\caption{Performance of Yin et al. method on each available test set in every phase. In each phase $i$, the accuracy for each previous test set, i.e., $Test_0$ to $Test_i$, is calculated.}\label{ Catastrophic Forgetting1}
 \scalebox{0.9}{
\begin{tabular}{|c|c|c|c|c|c|}
\hline
Test set& Phase 0 & Phase 1  & Phase 2 & Phase 3  & Phase 4 \\ \hline
$Test_0 $& 0.8956 &0.7939 &0.7988 &0.7226&0.5852               \\ \hline
$Test_1$&  -       &0.8458 &0.7566 &0.6556&0.6024                  \\\hline
$Test_2$& -&- &0.7547 &0.684 & 0.6025                   \\\hline
$Test_3$& -& -& -&0.8406 &0.7276                         \\\hline
$Test_4$&- &- &- &- &0.8318                       \\\hline
\end{tabular}
}
\end{table}
\begin{table}[!tbp]
\centering
\caption{Performance of our method method on each test set in every phase.  In each phase $i$, the accuracy for each previous test set, i.e., $Test_0$ to $Test_i$, is calculated.}\label{ Catastrophic Forgetting2}

 \scalebox{0.9}{
\begin{tabular}{|c|c|c|c|c|c|}
\hline
Test set& Phase 0 & Phase 1  & Phase 2 & Phase 3  & Phase 4 \\ \hline
$Test_0$ & 0.8956 &0.8876 &0.8822 &0.8756&0.8764               \\ \hline
$Test_1$&  -       &0.8462 &0.8382 &0.8287&0.8280                  \\\hline
$Test_2$& -&- &0.7711 &0.7502 & 0.7450                   \\\hline
$Test_3$& -& -& -&0.8448 &0.8359                         \\\hline
$Test_4$&- &- &- &- &0.8394                       \\\hline
\end{tabular}
}
\end{table}

\section{Conclusion}
In this paper, we propose and study the incremental KBQA task by focusing on learning from the evolving knowledge base. Considering the complexity of incremental KBQA, we design a new framework with emphasis on a novel margin-distilled loss and a semantic-based collaborative exemplar selection method. We re-organize the SimpleQuestion dataset to form an Incremental SimpleQuestion dataset to reveal the problem and evaluate the models. 

\bibliography{reference}

\begin{thebibliography}{30}
\providecommand{\natexlab}[1]{#1}
\providecommand{\url}[1]{\texttt{#1}}
\providecommand{\urlprefix}{URL }
\expandafter\ifx\csname urlstyle\endcsname\relax
  \providecommand{\doi}[1]{doi:\discretionary{}{}{}#1}\else
  \providecommand{\doi}{doi:\discretionary{}{}{}\begingroup
  \urlstyle{rm}\Url}\fi

\bibitem[{Bordes et~al.(2015)Bordes, Usunier, Chopra, and
  Weston}]{bordes2015large}
Bordes, A.; Usunier, N.; Chopra, S.; and Weston, J. 2015.
\newblock Large-scale simple question answering with memory networks.
\newblock \emph{arXiv preprint arXiv:1506.02075} .

\bibitem[{Castro et~al.(2018)Castro, Mar{\'\i}n-Jim{\'e}nez, Guil, Schmid, and
  Alahari}]{castro2018end}
Castro, F.~M.; Mar{\'\i}n-Jim{\'e}nez, M.~J.; Guil, N.; Schmid, C.; and
  Alahari, K. 2018.
\newblock End-to-end incremental learning.
\newblock In \emph{Proceedings of the European Conference on Computer Vision
  (ECCV)}, 233--248.

\bibitem[{Cauwenberghs and Poggio(2001)}]{cauwenberghs2001incremental}
Cauwenberghs, G.; and Poggio, T. 2001.
\newblock Incremental and decremental support vector machine learning.
\newblock In \emph{Advances in neural information processing systems},
  409--415.

\bibitem[{Dai, Li, and Xu(2016)}]{dai2016cfo}
Dai, Z.; Li, L.; and Xu, W. 2016.
\newblock CFO: Conditional Focused Neural Question Answering with Large-scale
  Knowledge Bases.
\newblock In \emph{Proceedings of the 54th Annual Meeting of the Association
  for Computational Linguistics (Volume 1: Long Papers)}, 800--810.

\bibitem[{Dong et~al.(2015)Dong, Wei, Zhou, and Xu}]{dong2015question}
Dong, L.; Wei, F.; Zhou, M.; and Xu, K. 2015.
\newblock Question answering over freebase with multi-column convolutional
  neural networks.
\newblock In \emph{Proceedings of the 53rd Annual Meeting of the Association
  for Computational Linguistics and the 7th International Joint Conference on
  Natural Language Processing (Volume 1: Long Papers)}, 260--269.

\bibitem[{Garcia-Duran, Duman{\v{c}}i{\'c}, and
  Niepert(2018)}]{garcia2018learning}
Garcia-Duran, A.; Duman{\v{c}}i{\'c}, S.; and Niepert, M. 2018.
\newblock Learning Sequence Encoders for Temporal Knowledge Graph Completion.
\newblock In \emph{Proceedings of the 2018 Conference on Empirical Methods in
  Natural Language Processing}, 4816--4821.

\bibitem[{Hao et~al.(2018)Hao, Liu, He, Liu, and Zhao}]{hao2018pattern}
Hao, Y.; Liu, H.; He, S.; Liu, K.; and Zhao, J. 2018.
\newblock Pattern-revising enhanced simple question answering over knowledge
  bases.
\newblock In \emph{Proceedings of the 27th International Conference on
  Computational Linguistics}, 3272--3282.

\bibitem[{He et~al.(2011)He, Chen, Li, and Xu}]{he2011incremental}
He, H.; Chen, S.; Li, K.; and Xu, X. 2011.
\newblock Incremental learning from stream data.
\newblock \emph{IEEE Transactions on Neural Networks} 22(12): 1901--1914.

\bibitem[{He and Golub(2016)}]{he2016character}
He, X.; and Golub, D. 2016.
\newblock Character-Level Question Answering with Attention.
\newblock In \emph{Proceedings of the 2016 Conference on Empirical Methods in
  Natural Language Processing}, 1598--1607.

\bibitem[{Hinton, Vinyals, and Dean(2015)}]{hinton2015distilling}
Hinton, G.; Vinyals, O.; and Dean, J. 2015.
\newblock Distilling the knowledge in a neural network.
\newblock \emph{arXiv preprint arXiv:1503.02531} .

\bibitem[{Hsiao, Huang, and Chen(2017)}]{hsiao2017integrating}
Hsiao, W.-C.; Huang, H.-H.; and Chen, H.-H. 2017.
\newblock Integrating subject, type, and property identification for simple
  question answering over knowledge base.
\newblock In \emph{Proceedings of the Eighth International Joint Conference on
  Natural Language Processing (Volume 1: Long Papers)}, 976--985.

\bibitem[{Lin, Socher, and Xiong(2018)}]{lin2018multi}
Lin, X.~V.; Socher, R.; and Xiong, C. 2018.
\newblock Multi-Hop Knowledge Graph Reasoning with Reward Shaping.
\newblock In \emph{Proceedings of the 2018 Conference on Empirical Methods in
  Natural Language Processing}, 3243--3253.

\bibitem[{Liu et~al.(2020)Liu, Su, Liu, Schiele, and Sun}]{liu2020mnemonics}
Liu, Y.; Su, Y.; Liu, A.-A.; Schiele, B.; and Sun, Q. 2020.
\newblock Mnemonics Training: Multi-Class Incremental Learning without
  Forgetting.
\newblock In \emph{Proceedings of the IEEE/CVF Conference on Computer Vision
  and Pattern Recognition}, 12245--12254.

\bibitem[{McCloskey and Cohen(1989)}]{mccloskey1989catastrophic}
McCloskey, M.; and Cohen, N.~J. 1989.
\newblock Catastrophic interference in connectionist networks: The sequential
  learning problem.
\newblock In \emph{Psychology of learning and motivation}, volume~24, 109--165.

\bibitem[{Mohammed, Shi, and Lin(2018)}]{mohammed2018strong}
Mohammed, S.; Shi, P.; and Lin, J. 2018.
\newblock Strong Baselines for Simple Question Answering over Knowledge Graphs
  with and without Neural Networks.
\newblock In \emph{Proceedings of the 2018 Conference of the North American
  Chapter of the Association for Computational Linguistics: Human Language
  Technologies, Volume 2 (Short Papers)}, 291--296.

\bibitem[{Rebuffi et~al.(2017)Rebuffi, Kolesnikov, Sperl, and
  Lampert}]{rebuffi2017icarl}
Rebuffi, S.-A.; Kolesnikov, A.; Sperl, G.; and Lampert, C.~H. 2017.
\newblock icarl: Incremental classifier and representation learning.
\newblock In \emph{Proceedings of the IEEE conference on Computer Vision and
  Pattern Recognition}, 2001--2010.

\bibitem[{Reddy, Lapata, and Steedman(2014)}]{reddy2014large}
Reddy, S.; Lapata, M.; and Steedman, M. 2014.
\newblock Large-scale semantic parsing without question-answer pairs.
\newblock \emph{Transactions of the Association for Computational Linguistics}
  2: 377--392.

\bibitem[{Shan et~al.(2020)Shan, Xu, Yang, Jia, and Xiang}]{shan2020learn}
Shan, G.; Xu, S.; Yang, L.; Jia, S.; and Xiang, Y. 2020.
\newblock Learn\#: A Novel Incremental Learning Method for Text Classification.
\newblock \emph{Expert Systems with Applications} 113198.

\bibitem[{Shin et~al.(2017)Shin, Lee, Kim, and Kim}]{shin2017continual}
Shin, H.; Lee, J.~K.; Kim, J.; and Kim, J. 2017.
\newblock Continual learning with deep generative replay.
\newblock In \emph{Advances in Neural Information Processing Systems},
  2990--2999.

\bibitem[{Suykens and Vandewalle(1999)}]{suykens1999least}
Suykens, J.~A.; and Vandewalle, J. 1999.
\newblock Least squares support vector machine classifiers.
\newblock \emph{Neural processing letters} 9(3): 293--300.

\bibitem[{Trivedi et~al.(2017)Trivedi, Dai, Wang, and Song}]{trivedi2017know}
Trivedi, R.; Dai, H.; Wang, Y.; and Song, L. 2017.
\newblock Know-evolve: deep temporal reasoning for dynamic knowledge graphs.
\newblock In \emph{Proceedings of the 34th International Conference on Machine
  Learning-Volume 70}, 3462--3471.

\bibitem[{Wang et~al.(2019)Wang, Zhang, Li, Hwang, Zong, and
  Li}]{wang2019incremental}
Wang, W.; Zhang, J.; Li, Q.; Hwang, M.-Y.; Zong, C.; and Li, Z. 2019.
\newblock Incremental Learning from Scratch for Task-Oriented Dialogue Systems.
\newblock In \emph{Proceedings of the 57th Annual Meeting of the Association
  for Computational Linguistics}, 3710--3720.

\bibitem[{Wang et~al.(2018)Wang, Zhang, Xu, and Mao}]{wang2018apva}
Wang, Y.; Zhang, R.; Xu, C.; and Mao, Y. 2018.
\newblock The APVA-TURBO approach to question answering in knowledge base.
\newblock In \emph{Proceedings of the 27th International Conference on
  Computational Linguistics}, 1998--2009.

\bibitem[{Wu et~al.(2019)Wu, Huang, Weng, Zheng, Zhang, Yan, and
  Chen}]{wu2019learning}
Wu, P.; Huang, S.; Weng, R.; Zheng, Z.; Zhang, J.; Yan, X.; and Chen, J. 2019.
\newblock Learning Representation Mapping for Relation Detection in Knowledge
  Base Question Answering.
\newblock In \emph{Proceedings of the 57th Annual Meeting of the Association
  for Computational Linguistics}, 6130--6139.

\bibitem[{Xu et~al.(2018)Xu, Wu, Wang, Yu, Chen, and
  Sheinin}]{xu2018exploiting}
Xu, K.; Wu, L.; Wang, Z.; Yu, M.; Chen, L.; and Sheinin, V. 2018.
\newblock Exploiting Rich Syntactic Information for Semantic Parsing with
  Graph-to-Sequence Model.
\newblock In \emph{Proceedings of the 2018 Conference on Empirical Methods in
  Natural Language Processing}, 918--924.

\bibitem[{Yih, He, and Meek(2014)}]{yih2014semantic}
Yih, W.-t.; He, X.; and Meek, C. 2014.
\newblock Semantic parsing for single-relation question answering.
\newblock In \emph{Proceedings of the 52nd Annual Meeting of the Association
  for Computational Linguistics (Volume 2: Short Papers)}, 643--648.

\bibitem[{Yin et~al.(2016)Yin, Yu, Xiang, Zhou, and
  Sch{\"u}tze}]{yin2016simple}
Yin, W.; Yu, M.; Xiang, B.; Zhou, B.; and Sch{\"u}tze, H. 2016.
\newblock Simple Question Answering by Attentive Convolutional Neural Network.
\newblock In \emph{Proceedings of COLING 2016, the 26th International
  Conference on Computational Linguistics: Technical Papers}, 1746--1756.

\bibitem[{Zhang et~al.(2018)Zhang, Dai, Kozareva, Smola, and
  Song}]{zhang2018variational}
Zhang, Y.; Dai, H.; Kozareva, Z.; Smola, A.~J.; and Song, L. 2018.
\newblock Variational reasoning for question answering with knowledge graph.
\newblock In \emph{Thirty-Second AAAI Conference on Artificial Intelligence}.

\bibitem[{Zhang et~al.(2016)Zhang, Liu, He, Ji, Liu, Wu, and
  Zhao}]{zhang2016question}
Zhang, Y.; Liu, K.; He, S.; Ji, G.; Liu, Z.; Wu, H.; and Zhao, J. 2016.
\newblock Question answering over knowledge base with neural attention
  combining global knowledge information.
\newblock \emph{arXiv preprint arXiv:1606.00979} .

\bibitem[{Zhu et~al.(2018)Zhu, Ikeda, Pang, Ban, and
  Sarrafzadeh}]{zhu2018merging}
Zhu, L.; Ikeda, K.; Pang, S.; Ban, T.; and Sarrafzadeh, A. 2018.
\newblock Merging weighted SVMs for parallel incremental learning.
\newblock \emph{Neural Networks} 100: 25--38.

\end{thebibliography}
\end{document}